\documentclass[]{elsarticle}

\makeatletter
\def\ps@pprintTitle{%
	\let\@oddhead\@empty
	\let\@evenhead\@empty
	\def\@oddfoot{\footnotesize\itshape
	\parbox{\textwidth}{
	\textcopyright 2020. This manuscript version is made available under the CC-BY-NC-ND 4.0 license\\ \url{http://creativecommons.org/licenses/by-nc-nd/4.0/}
}
}%
	\let\@evenfoot\@oddfoot}
\makeatother

\usepackage[hidelinks]{hyperref}

\journal{International Journal of Forecasting}

\biboptions{authoryear}

\usepackage{amsmath}
\usepackage{amsfonts}

\usepackage[table,xcdraw]{xcolor} % for table
\usepackage{multirow} % for table
\usepackage{pgfplots} % for plots
\usetikzlibrary{shapes,arrows} % for flowcharts
\usetikzlibrary{external}
\usepackage{subfig} % for subfigures
\usepackage[utf8]{inputenc} % for Estonian characters

\usepackage{xcolor}
\usepackage{lipsum} % dummy text for the MWE

\usepackage{enumerate}

\newcommand{\anonymize}[1]{#1}

\begin{document}
	
This is an Accepted Manuscript version of an article accepted for publication/published in Internation Journal of Forecasting. The Version of Record is available online at \url{https://doi.org/10.1016/j.ijforecast.2019.02.018}.

 \begin{frontmatter}

 \title{Correlated daily time series and forecasting in the M4 competition}

 %% Group authors per affiliation:
 \author[mymainaddress]{Anti Ingel\corref{mycorrespondingauthor}}
 \ead{anti.ingel@ut.ee}
 \author[mymainaddress]{Novin Shahroudi}
 \ead{novin@ut.ee}
 \author[mymainaddress]{Markus Kängsepp}
 \ead{markus93@ut.ee}
 \author[mymainaddress]{Andre Tättar}
 \ead{andre.tattar@ut.ee}
 \author[mymainaddress]{Viacheslav Komisarenko}
 \ead{viacheslav.komisarenko@ut.ee}
 \author[mymainaddress]{Meelis Kull}
 \ead{meelis.kull@ut.ee}

 \cortext[mycorrespondingauthor]{Corresponding author}

 \address[mymainaddress]{
 Institute of Computer Science, University of Tartu, J. Liivi 2, Tartu 50409, Estonia
 }

 \begin{abstract}
 We participated in the M4 competition for time series forecasting and describe here our methods for forecasting daily time series. 
 We used an ensemble of five statistical forecasting methods and a method that we refer to as the correlator. 
 Our retrospective analysis using the ground truth values published by the M4 organisers after the competition demonstrates that the correlator was responsible for most of our gains over the naive constant forecasting method.
 We identify data leakage as one reason for its success, partly due to test data selected from different time intervals, and partly due to quality issues in the original time series. We suggest that future forecasting competitions should provide actual dates for the time series so that some of those leakages could be avoided by the participants.
 \end{abstract}


 \end{frontmatter}
%\footnotetext{\textcopyright 2020. This manuscript version is made available under the CC-BY-NC-ND 4.0 license \url{http://creativecommons.org/licenses/by-nc-nd/4.0/}}
% \linenumbers

\section{Introduction}

The task in the M4 time series forecasting competition was to provide forecasts for 100,000 numeric time series with different frequencies and origins~\citep{m4}. Participants of the competition were free to choose their methods differently depending on the type of time series and in the following we will describe how we compiled our forecasts on the 4227 time series from the Daily category. Each of those time series contained a numeric value for each day over a time interval ranging from 93 to 9919 days, and the task was to forecast the values for the next 14 days.

We participated in the M4 competition as a team of \anonymize{24 students and 1 instructor and our submission was created in a dedicated M4 seminar at the University of Tartu}. We formed subteams to try out different approaches and evaluated these in an internal competition where we withheld the last available 14 days as an internal validation set. Our final submission was obtained by choosing a different ensemble of models for each category of time series: Hourly, Daily, Weekly, Monthly, Quarterly, Yearly. Forecasting the Daily time series was done using an ensemble of 5 statistical methods and one correlation-based method, which we refer to as the \emph{correlator}. After the organisers of the M4 competition published the full time series we performed an analysis revealing that the key to the good performance of our submission was the correlator.

In Section~\ref{sec:methods} we will describe our methods used for forecasting the Daily time series in the M4 competition. 
In Section~\ref{sec:results} we evaluate the methods separately on the official M4 test data and demonstrate the role of the correlator in the final performance of our submission. 
In Section~\ref{sec:discussion} we discuss the reasons behind the success of the correlator and its implications for forecasting competitions.

\section{Methods} \label{sec:methods}

Our forecasts for the Daily time series in the M4 competition were obtained with two methods. For a subset of the 4227 time series a forecasting method which we refer to as the \emph{correlator} was used. This method compares the time series in a given dataset between themselves to find highly correlated segments  and uses this information to produce the forecasts. The method is used only on those time series for which the correlation is larger than a given threshold. The method is described in detail in Section~\ref{sec:correlator}.

On the rest of the time series, that is for the time series for which the correlator did not find high correlations, an ensemble of five models was used to produce the forecasts. The ensemble is described in detail in Section~\ref{sec:model_fitting}. 

The methods were selected based on their performance on the holdout validation dataset that we had separated from the training data. From each training time series, we removed the last 14 time points and these removed values formed the validation data. After this, we were able to evaluate any method by calculating its performance measure on the validation set. 

As a post-processing step, all the negative predicted values were replaced by zeros. This was done because the original dataset did not contain any negative values and thus avoiding forecasting negative values was deemed beneficial.

\subsection{Ensembling model}\label{sec:model_fitting}

Our ensembling method used five models:
\begin{itemize}
    \item Naive model, which makes a constant forecast equal to the last observed value in the time series;
    \item an ETS model (see \ref{sec:arima} for details);
    \item two ARIMA models (see \ref{sec:arima} for details);
    \item a custom model based on time series decomposition (see \ref{sec:fearless} for details).
\end{itemize}
These five methods were chosen based on empirical results on the holdout validation data. In addition we considered Holt-Winters~\citep{ETS}, BATS~\citep{bats}, LSTM networks~\citep{lstm} and a combination of ARIMA and a neural network~\citep{arima_nn}. 
On each time series, we trained each of the chosen methods to obtain five models. The ensemble forecast was obtained by taking the median of the five single method based forecasts for each time point in the forecasting horizon. Flowchart of the method can be seen in Figure~\ref{fig:flowchart}.

\subsection{Correlator}\label{sec:correlator}

The idea behind the correlator is to look for highly correlating patterns between different time series at different times. If such a correlation is found between the last 14 days of the current time series and some other 14 day period that continues for at least 14 days in the training set, then the correlator uses these subsequent 14 days to produce a forecast for the current series. The practical applications of this idea are limited, because it relies on the existence of correlating patterns between different times in different time series. However, as we show and discuss in the following sections, this idea worked surprisingly well in the M4 competition.

\begin{figure}[h]
	\begin{minipage}{.5\textwidth}
		\centering
		\tikzstyle{decision} = [diamond, draw, fill=blue!20,
    text width=5em, text badly centered, node distance=2.5cm, inner sep=0pt]
\tikzstyle{block} = [rectangle, draw, fill=blue!20,
    text width=10em, text centered, rounded corners, minimum height=4em]
\tikzstyle{line} = [draw, very thick, color=black!50, -latex']
\tikzstyle{cloud} = [draw, ellipse,fill=red!20,
    minimum height=2em]

\newcommand*{\ArrowLength}{2.0em}
\newcommand*{\MyRightArrow}[1][]{
    \tikz [-stealth, red, yshift=0.5ex, baseline] 
        \draw [-stealth, #1] (0,0) -- (\ArrowLength,0) ;
}

\begin{tikzpicture}[scale=0.6, every node/.style={scale=0.6}, node distance = 2.5cm, auto]
	% Nodes
	\node [cloud] (start) {Start};
	\node [block, right of=start, node distance=3.3cm] (callcorr) {Call correlator\\(See Figure~\ref{fig:corr_flow})};
	\node [decision, below of=callcorr, node distance=3.1cm] (corr) {Correlator returned forecast};
	\coordinate [below of=corr] (dummy);
	\node [block, right of=dummy] (corryes) {Use correlator prediction};
	\node [block, left of=dummy] (corrno) {For each time point predict median of 5 model predictions};
	\node [block, below of=dummy] (post) {Replace negative predictions with 0};
	\node [cloud, below of=post, node distance=2cm] (stop) {Stop};

	% Arrows
	\path [line] (start) -- (callcorr);
	\path [line] (corr) -- node [midway, above, sloped, color=black] {Yes} (corryes);
	\path [line] (corr) -- node [midway, above, sloped, color=black] {No} (corrno);
	\path [line] (corrno) -- (post);
	\path [line] (corryes) -- (post);
	\path [line] (post) -- (stop);
	\path[line] (callcorr) -- (corr);
\end{tikzpicture}
		\caption{Flowchart of prediction process}
		\label{fig:flowchart}
	\end{minipage}
	\begin{minipage}{.5\textwidth}
		\centering
		\tikzstyle{decision} = [diamond, draw, fill=blue!20,
    text width=4.5em, text badly centered, node distance=2.5cm, inner sep=0pt]
\tikzstyle{block} = [rectangle, draw, fill=blue!20,
    text width=10em, text centered, rounded corners, minimum height=4em]
\tikzstyle{line} = [draw, very thick, color=black!50, -latex']
\tikzstyle{cloud} = [draw, ellipse,fill=red!20,
    minimum height=2em]

\newcommand*{\ArrowLength}{2.0em}
\newcommand*{\MyRightArrow}[1][]{
    \tikz [-stealth, red, yshift=0.5ex, baseline] 
        \draw [-stealth, #1] (0,0) -- (\ArrowLength,0) ;
}

\begin{tikzpicture}[scale=0.6, every node/.style={scale=0.6}, node distance = 2cm, auto]
	% Nodes
	\node [cloud] (start) {Start};
	\node [block, right of=start, node distance=3.3cm] (cross) {Calculate correlations between segments};
	\node [block, below of=cross] (max) {Find segment with maximum correlation};
	\node [decision, below of=max] (cond1) {Inequality \eqref{eq:condition2} holds?};
	\coordinate [below of=cond1] (dummy);
	\node [block, right of=dummy] (cond1yes) {Scale the potential forecast};
	\node [block, left of=dummy] (cond1no) {Correlator will not be used};
	\node [cloud, below of=cond1no] (ensemble) {Return nothing};
	\node [decision, below of=cond1yes] (cond2) {Inequality \eqref{eq:condition1} holds?};
	\coordinate [below of=cond2] (dummy2);
	%\node [block, left of=dummy2] (cond2yes) {Use scaled segment as forecast};
	\node [cloud, left of=dummy2] (cond2yes) {Return forecast};
	\node [block, right of=dummy2] (cond2no) {Find next largest correlation};
	%\node [cloud, left of=cond2yes, node distance=3cm] (stop) {Stop};

	% Arrows
	\path [line] (start) -- (cross);
	\path [line] (cross) -- (max);
	\path [line] (max) -- (cond1);
	\path [line] (cond1) -- node [midway, above, sloped, color=black] {Yes} (cond1yes);
	\path [line] (cond1) -- node [midway, above, sloped, color=black] {No} (cond1no);
	\path [line] (cond1yes) -- (cond2);
	\path [line] (cond2) -- node [midway, above, sloped, color=black] {Yes} (cond2yes);
	\path [line] (cond2) -- node [midway, above, sloped, color=black] {No} (cond2no);
	%\path [line] (cond2yes) -- (stop);
	\path [line] (cond1no) -- (ensemble);
	\path [line] (cond2no) to[in=10] (cond1);
	%\path [line] (ensemble) -- (stop);

\end{tikzpicture}
		\caption{Flowchart of the correlator.}
		\label{fig:corr_flow}
	\end{minipage}
\end{figure}

There are 4227 time series in the Daily category with varying lengths and starting times. Let us denote the $j$-th time series of the Daily dataset as $y^j$ and its data points as $y^j_1,y^j_2,\dots,y^j_{n_j}$ where $n_j$ denotes the length of the $j$-th time series. Further, for convenience we will use notation $[i_1:i_2]$ for interval indexing, that is we denote data points of time series $y^j$ from index $i_1$ until index $i_2$ as $y^j_{[i_1:i_2]}:=(y^j_{i_1},\dots,y^j_{i_2})$.

To produce forecasts for the $j$-th time series $y^j_1,\dots,y^j_{n_j}$, the correlator first takes its last 14 values $y^j_{[n_j-13:n_j]}$ and calculates Pearson correlation coefficient between this vector of 14 values and all the non-terminal segments of the same length across all time-series, where by non-terminal we mean segments which have at least 14 subsequent data points. We denote the Pearson correlation coefficient between the end segment of the $j$-th time series and a segment of the $k$-th time series $y^k_1,\dots,y^k_{n_k}$ ending at $\tau\in\{14,\dots,n_k-14\}$ as follows
\begin{align}
    r_{jk}(\tau):=\mbox{pearson}\Big(y^j_{[n_j-13\colon n_j]}, y^k_{[\tau-13\colon\tau]}\Big),
\end{align}
where for any length $n$ and for any vectors $a=(a_1,\dots,a_n)$ and $b=(b_1,\dots,b_n)$ of real numbers, $\mbox{pearson}(a,b)$ is the Pearson correlation coefficient
\begin{align}
    \mbox{pearson}(a,b)&=\frac{\sum_{i=1}^n(a_i-\mbox{mean}(a))(b_i-\mbox{mean}(b))}{\mbox{std}(a)\cdot\mbox{std}(b)},
\end{align}
where
\begin{align}
    \mbox{mean}\big(a\big)=\frac{1}{n}\sum_{i=1}^na_i,\qquad
    \mbox{std}\big(a\big)=\sqrt{\frac{1}{n}\sum_{i=1}^n(a_i-\mbox{mean}(a))^2}.
\end{align}

Next, we find for each time series $y^j_1,\dots,y^j_{n_j}$ the values $k$ and $\tau$ which maximise $r_{jk}(\tau)$. In other words, the end segment of a time series $y^j$ is compared to all given time series and a time series $y^k$ is found with a segment that is most similar to the end segment of time series $y^j$ and continues for at least 14 data points. The next 14 data points $y^k_{[\tau+1:\tau+14]}$ from time series $y^k$ are then scaled and translated using a function $f$ defined as
\begin{align}
    f(x)=\left(x-\mbox{mean}\left(y^k_{[\tau-13:\tau]}\right)\right)\frac{\mbox{std}\left(y^j_{[n_j-13:n_j]}\right)}{\mbox{std}\left(y^k_{[\tau-13:\tau]}\right)}+\mbox{mean}\left(y^j_{[n_j-13:n_j]}\right),
\end{align}
where $x$ is a real number. Function $f$ maps values $y^k_{[\tau-13:\tau]}$ approximately to values $y^j_{[n_j-13:n_j]}$. The obtained result is considered as the forecast for the time series $y^j$. We denote this result as  $\hat{y}^j_{n_j+1}:=f(y^k_{\tau+1}),\dots,\hat{y}^j_{n_j+14}:=f(y^k_{\tau+14})$.

The resulting forecast from the correlator is used as the actual forecast only if two conditions are met.
The first condition that must hold is that the correlation must be at least 0.9999, that is
\begin{align}\label{eq:condition2}
r_{jk}(\tau)\geq 0.9999.
\end{align}
This condition was used to avoid spurious correlations. As the segment over which the correlation is calculated is only 14 data points long, a high threshold had to be used. 

The second condition is 
\begin{align}\label{eq:condition1}
    \mbox{std}\left(\hat{y}^j_{[n_j+1:n_j+14]}\right)\leq 2.5\cdot\mbox{std}\left(y^j_{[n_j-13:n_j]}\right)
% 	\hat{s}_{j,n+1}\leq 2.5 \cdot s_{j,n-13}
\end{align}
which means that the standard deviation of the forecast $\hat{y}^j_{[n_j+1:n_j+14]}$ must not exceed 2.5 times the standard deviation of the end segment of the time series $y^j$. Without this condition, the forecast can have a sudden change in standard deviation compared to the end segment of the time series, e.g. the forecast can contain large jumps. Therefore, the correlator skips forecasts that do not satisfy Equation~\eqref{eq:condition1} to find forecasts more similar to the end segment of current time series. 

If the Equation~\eqref{eq:condition1} does not hold, new values $k'$ and $\tau'$ are found such that they maximise $r_{jk'}(\tau')$ and differ from the previously found $k$ and $\tau$. This is iterated until suitable values are found or no suitable values exist, in which case the correlator is not used. The flowchart of the correlator can be seen in Figure~\ref{fig:corr_flow}.

The only changeable parameters for this method are the number of values used from the end of the time series, in this work 14, the correlation cut-off threshold in \eqref{eq:condition2} and the multiplier for standard deviation comparison in \eqref{eq:condition1}. All of these were chosen in this work based on empirical results on the validation dataset. However, our retrospective analysis showed that the threshold of 0.9999 was not optimal and Equation~\eqref{eq:condition1} was not needed (see Table~\ref{tab:std}).

\subsection{Differences from submission}\label{sec:bugs}

After the submission of our forecasts to the M4 competition, we found two bugs in our implementation of the correlator. Therefore, the method used for producing the submission has two differences from the description given in Section \ref{sec:correlator}.

First, in the original submission, the correlator was used only on time series D1-D2138 and not used on any of the time series D2139-D4227. This was probably caused by an unseen memory error when applying the correlator. We will refer to this error as Bug 1.

The second difference is that instead of Equation \eqref{eq:condition1}, due to human error the condition
\begin{align}
	\mbox{std}\left(\hat{y}^j_{[n_j+1:n_j+14]}\right)\leq 2.5\cdot \mbox{std}\left(y^k_{[\tau-13:\tau]}\right)
\end{align}
was used instead. We will refer to this error as Bug 2.

\section{Results}\label{sec:results}

In this section we present the performance of all models presented in Section~\ref{sec:model_fitting} individually and compare these to the performances of the ensemble with and without the correlator to determine which part of our method was most successful. We also compare the performance of the correlator with different parameter values. Tables \ref{tab:single_models}, \ref{tab:std} and \ref{tab:categories} show the post-submission analysis of the data, using the actual test set that was made public by the organisers after the competition.

In our experiments, we use the correlator's parameter values mentioned in Section~\ref{sec:correlator} if not stated otherwise. The results were evaluated using MASE, sMAPE and the overall weighted average (OWA) of the relative MASE and the relative sMAPE~\citep{comp_guide}. Relative MASE and relative sMAPE are obtained by dividing MASE and sMAPE of the forecast by MASE and sMAPE of naive forecast respectively, and OWA is the mean of the obtained values. Therefore, OWA of less than 1 is better than Naive and more than one is worse than Naive. In the following, the presented values of MASE, sMAPE and OWA are the average values over the considered forecasts.

\begin{table}[h]
	\def\arraystretch{1.5}
	\caption{OWA for single models, ensemble and benchmark method Comb on Daily dataset. Benchmark method Comb is the arithmetic average of the Simple, Holt and Damped exponential smoothing models that is also used by~\cite{m4}. We evaluated the performance of Comb using its implementation available in the M4 code repository\protect\footnotemark.}
	\label{tab:single_models}
	\resizebox{\textwidth}{!}
	{
\begin{tabular}{|l|l|l|l|l|l|l|l|l|}
\hline
\multicolumn{2}{|l|}{\textbf{}}                                                                        & \textbf{Macro}                & \textbf{Micro}                & \textbf{Demographic}          & \textbf{Industry}             & \textbf{Finance}              & \textbf{Other}                & \textbf{All}                  \\ \hline
                                        & \cellcolor[HTML]{EFEFEF}\textbf{ARIMA 1}                     & \cellcolor[HTML]{EFEFEF}1.160 & \cellcolor[HTML]{EFEFEF}1.023 & \cellcolor[HTML]{EFEFEF}1.142 & \cellcolor[HTML]{EFEFEF}1.016 & \cellcolor[HTML]{EFEFEF}1.052 & \cellcolor[HTML]{EFEFEF}1.038 & \cellcolor[HTML]{EFEFEF}1.041 \\ \cline{2-9} 
                                        & \textbf{ARIMA 2}                                             & 1.123                         & 1.009                         & 1.144                         & 1.031                         & 1.035                         & 1.051                         & 1.033                         \\ \cline{2-9} 
                                        & \cellcolor[HTML]{EFEFEF}\textbf{ETS}                         & \cellcolor[HTML]{EFEFEF}1.022 & \cellcolor[HTML]{EFEFEF}0.989 & \cellcolor[HTML]{EFEFEF}0.983 & \cellcolor[HTML]{EFEFEF}1.012 & \cellcolor[HTML]{EFEFEF}0.991 & \cellcolor[HTML]{EFEFEF}1.006 & \cellcolor[HTML]{EFEFEF}0.996 \\ \cline{2-9} 
                                        & \textbf{Custom method}                                       & 1.336                         & 1.360                         & 1.050                         & 1.342                         & 1.380                         & 1.304                         & 1.354                         \\ \cline{2-9} 
\multirow{-5}{*}{\textbf{Single model}} & \cellcolor[HTML]{EFEFEF}\textbf{Naive}                       & \cellcolor[HTML]{EFEFEF}1.000 & \cellcolor[HTML]{EFEFEF}1.000 & \cellcolor[HTML]{EFEFEF}1.000 & \cellcolor[HTML]{EFEFEF}1.000 & \cellcolor[HTML]{EFEFEF}1.000 & \cellcolor[HTML]{EFEFEF}1.000 & \cellcolor[HTML]{EFEFEF}1.000 \\ \hline
                                        & \textbf{No correlator}                                       & 1.020                         & 0.990                         & 0.999                         & 1.008                         & 0.987                         & 1.011                         & 0.995                         \\ \cline{2-9} 
                                        & \cellcolor[HTML]{EFEFEF}\textbf{Correlator with Bug 1 and 2} & \cellcolor[HTML]{EFEFEF}0.805 & \cellcolor[HTML]{EFEFEF}0.785 & \cellcolor[HTML]{EFEFEF}0.849 & \cellcolor[HTML]{EFEFEF}1.007 & \cellcolor[HTML]{EFEFEF}0.985 & \cellcolor[HTML]{EFEFEF}1.011 & \cellcolor[HTML]{EFEFEF}0.930 \\ \cline{2-9} 
                                        & \textbf{Correlator with Bug 2}                               & 0.805                         & 0.785                         & 0.849                         & 1.007                         & 0.944                         & 0.981                         & 0.908                         \\ \cline{2-9} 
\multirow{-4}{*}{\textbf{Ensemble}}     & \cellcolor[HTML]{EFEFEF}\textbf{Correlator}                  & \cellcolor[HTML]{EFEFEF}0.776 & \cellcolor[HTML]{EFEFEF}0.778 & \cellcolor[HTML]{EFEFEF}0.814 & \cellcolor[HTML]{EFEFEF}1.007 & \cellcolor[HTML]{EFEFEF}0.939 & \cellcolor[HTML]{EFEFEF}0.979 & \cellcolor[HTML]{EFEFEF}0.903 \\ \hline
\textbf{Benchmark}                      & \textbf{Comb}                                                & 1.015                         & 0.980                         & 0.978                         & 1.005                         & 0.958                         & 1.005                         & 0.979                         \\ \hline
\end{tabular}}
\end{table}
\footnotetext{\url{https://github.com/M4Competition/M4-methods}}
\begin{table}[h]
\def\arraystretch{1.5}
\caption{Comparison of correlator's performance on Daily dataset for different values of STD ratio threshold in Equation~\eqref{eq:condition1}, dash means that Equation~\eqref{eq:condition1} was not used. Correlator MASE, sMAPE and OWA are calculated from the forecasts of the correlator, that is only subset of Daily dataset is considered. Full OWA is calculated from the full dataset using forecasts obtained from the combination of ensemble and correlator.}
\label{tab:std}
\resizebox{\textwidth}{!}
{
\begin{tabular}{|l|l|l|l|l|l|l|l|l|l|l|l|l|}
\hline
                              & \multicolumn{4}{l|}{\textbf{Correlation threshold 0.9999}} & \multicolumn{4}{l|}{\textbf{Correlation threshold 0.999}} & \multicolumn{4}{l|}{\textbf{Correlation threshold 0.99}} \\ \hline
\textbf{STD ratio threshold}  & \textbf{2}    & \textbf{2.5}    & \textbf{3}    & -        & \textbf{2}    & \textbf{2.5}    & \textbf{3}    & -       & \textbf{2}    & \textbf{2.5}    & \textbf{3}   & -       \\ \hline
\rowcolor[HTML]{EFEFEF} 
\textbf{Correlator used (\#)} & 480           & 518             & 541           & 566      & 743           & 794             & 827           & 855     & 1704          & 1772            & 1827         & 1867    \\ \hline
\textbf{Correlator used (\%)} & 11.4          & 12.3            & 12.8          & 13.4     & 17.6          & 18.8            & 19.6          & 20.2    & 40.3          & 41.9            & 43.2         & 44.2    \\ \hline
\rowcolor[HTML]{EFEFEF} 
\textbf{Correlator MASE}                 & 0.191         & 0.183           & 0.196         & 0.254    & 0.227         & 0.214           & 0.221         & 0.275   & 2.176         & 2.134           & 2.140        & 2.143   \\ \hline
\textbf{Correlator sMAPE}                & 0.310         & 0.298           & 0.334         & 0.386    & 0.299         & 0.284           & 0.306         & 0.356   & 2.485         & 2.411           & 2.411        & 2.408   \\ \hline
\rowcolor[HTML]{EFEFEF} 
\textbf{Correlator OWA}                  & 0.098         & 0.092           & 0.098         & 0.115    & 0.109         & 0.101           & 0.104         & 0.121   & 0.729         & 0.709           & 0.704        & 0.699   \\ \hline
\textbf{Full OWA}             & 0.913         & 0.903           & 0.897         & 0.891    & 0.875         & 0.863           & 0.854         & 0.848   & 0.890         & 0.876           & 0.870        & 0.864   \\ \hline
\end{tabular}}
\end{table}

\begin{table}[h]
    \def\arraystretch{1.5}
    \caption{Performance of correlator on all categories. Correlator MASE, sMAPE and OWA are calculated from the forecasts of the correlator, that is only subset of Daily dataset is considered. Correlator used 14-time-point window for calculating correlations for all categories except for Yearly, for which 13 was used, because some of the time series in Yearly dataset had only 13 data points.}
	\label{tab:categories}
	\resizebox{\textwidth}{!}
	{
\begin{tabular}{|l|l|l|l|l|l|l|}
\hline
                              & \textbf{Hourly} & \textbf{Daily} & \textbf{Weekly} & \textbf{Monthly} & \textbf{Quarterly} & \textbf{Yearly} \\ \hline
\rowcolor[HTML]{EFEFEF} 
\textbf{Time series (\#)}     & 414             & 4227           & 359             & 48000            & 24000              & 23000           \\ \hline
\textbf{Correlator used (\#)} & 25              & 518            & 7               & 1357             & 450                & 178             \\ \hline
\rowcolor[HTML]{EFEFEF} 
\textbf{Correlator used (\%)} & 6.0             & 12.3           & 1.9             & 2.8              & 1.9                & 0.8             \\ \hline
\textbf{Correlator MASE}                 & 0.124           & 0.183          & 1.732           & 4.972            & 0.552              & 2.929           \\ \hline
\rowcolor[HTML]{EFEFEF} 
\textbf{Correlator sMAPE}                & 0.560           & 0.298          & 0.632           & 8.144            & 1.396              & 5.321           \\ \hline
\textbf{Correlator OWA}                  & 0.028           & 0.092          & 0.360           & 0.580            & 0.128              & 0.444           \\ \hline
\end{tabular}}
\end{table}

As can be seen from Table~\ref{tab:single_models}, the main improvement over Naive was achieved thanks to using the correlator. From single models, only ETS achieved better performance than Naive with OWA of 0.996. The ensemble of five models only slightly improved the result to 0.995. As can be seen in Table~\ref{tab:std}, by using the correlator, it would have been possible to obtain an OWA of 0.863. However, our submission achieved only 0.930 due to using a higher correlation threshold (0.9999 instead of 0.999) and the presence of two bugs in the code, as described in Section~\ref{sec:bugs}. Additional analysis showed that using only one of the two ARIMA models resulted in better OWA: 0.980 when only ARIMA 1 was used and 0.982 when only ARIMA 2 was used.

From Table~\ref{tab:std} it can be seen that using the value of 2.5 for the constant in Equation~\eqref{eq:condition1} gives the smallest OWA for the correlator's forecasts (see the row 'Correlator OWA' in the table). However, the OWA for the combination of ensemble and correlator is the smallest when Equation \eqref{eq:condition1} is not used at all (see the row 'Full OWA'). Without Equation \eqref{eq:condition1} an OWA of 0.848 could have been achieved by the proposed method.

Finally, Table~\ref{tab:categories} shows the performance of the correlator on other categories of the M4 dataset. We can see that the correlator was the most successful on Daily and Hourly datasets where it was used on 12\% and 6\% of the time series respectively and achieved OWA of less than 0.1. On the rest of the categories, the correlator was used on less than 3\% of time series and in most cases OWA was considerably larger than on Daily and Hourly datasets. This suggests that the similarities between time series were most exploitable in Daily and Hourly category.

\section{Discussion}\label{sec:discussion}

As can be seen from Table~\ref{tab:single_models}, most of the improvement over Naive was achieved thanks to using the correlator. This implies that the Daily dataset contained time series that had highly correlated short segments between each other. However, it turns out that the correlated regions are actually often much longer. To understand the nature of correlations in this dataset better, we performed the following further analysis. 

Similarly to the correlator of Section 2.2, we cross-correlated each time series $y^j$ with every other time series $y^k$ by sliding one time series along the other. However, instead of calculating the Pearson correlation coefficient after each shift between 14 data points we calculated it globally on the whole overlapping region. Formally, this means that for every $\tau\in\{14,\dots,n_k-14\}$ we calculated the following correlation:
\begin{align}\label{eq:new_cor}
    r'_{jk}(\tau):=\mbox{pearson}\left(y^j_{[n_j-(\min(n_j,\tau)-1):n_j]}, y^k_{[\tau-(\min(n_j,\tau)-1):\tau]}\right).
\end{align}
As in Section 2.2, we found for each time series $y^j$ the values $k$ and $\tau$ which maximise $r'_{jk}(\tau)$. As a result, for 1083 time series we found $k$ and $\tau$ with $r'_{jk}\geq 0.995$. 
Detailed inspection revealed that some of those high correlations were due to a single big jump in both time series which after alignment resulted in a high correlation, even though the regions before the jump and after the jump were not correlated, respectively. We discarded such cases with jumps manually (the detailed list is available in the code repository, see Appendix A). 

The remaining set $C$ of 1004 pairs of highly correlated time series $(y^j,y^k)$ with the respective shifts $\tau$ constitute cases with global correlations, not just a short correlated segment. Note that the length of the overlapping segment between time series $y^j$ and $y^k$ in \eqref{eq:new_cor} is $\min(n_j,\tau)$. Figure~\ref{fig:histogram} shows the histogram over the overlap lengths $\min(n_j,\tau)$ of the pairs in set $C$. We can see that the Daily dataset even contains time series that have highly correlated segments with length up to around 4000 data points. However, highly correlated segments of less than 1000 data points are the most frequent. In the following we split the set $C$ of 1004 cases into 4 categories.

\begin{enumerate}[T1:]
    \item {\bf Self-correlations} (cases where $(y^j,y^j)\in C$). These are time series where the end is highly correlated with the beginning. While short self-correlations could be explained by periodicity, our identified cases can have very long repeats, potentially pointing at problems in data, for an example see Figure~\ref{fig:match_itself}. 
    \item {\bf Mutual correlations} (cases where $(y^j,y^k)\in C$ and $(y^k,y^j)\in C$ for $j\neq k$). These are pairs of time series where the end of one series correlates with the beginning of the other, and vice versa. For an example, see Figure~\ref{fig:t2_example}.
    \item {\bf Synchronised correlations} (cases where $j\neq k$ and the correlated region corresponds to the same dates in both time series). In such case using the end of $y^k$ to forecast the continuation of $y^j$ would be using data from the future (forecast earlier dates using data from later dates), which is of course not doable in practical applications and demonstrates data leakage~\citep{data_leakage} in the M4 data. For an example, see Figure~\ref{fig:scaled_match}.
At the time of competition we could not identify these cases, as the actual starting dates of time series were only published by the M4 organisers after the competition~\citep{m4}. 
    \item {\bf Unsynchronised correlations} (cases where $j\neq k$ and the correlated region corresponds to different dates in both time series). Depending on which time series is earlier, this might be a leakage or not. The example in Figure~\ref{fig:match_shift} shows series that are claimed to start at the same date but become almost perfectly correlated after shifting one of them, possibly suggesting that the start date information might not be correct. 
\end{enumerate}
 
Table~\ref{tab:t_cat} shows the number of time series pairs in each of the categories T1, T2, T3 and T4. Note that we made the categories mutually exclusive by excluding those that belong to narrower categories T1 and T2 from the wider categories T3 and T4.

Note that above we have only categorised correlations where one of the segments is at the end and another at the beginning of some time series. Additionally, we identified other more peculiar kinds of correlations, such as the example shown in Figure~\ref{fig:correlated_new}, where one time series can be split into 3 large segments and rearranged almost perfectly into one another time series.

\begin{figure}[h]
    \hspace{-1cm}\begin{minipage}{.6\textwidth}
    \centering
    \resizebox{5cm}{!}{% This file was created by matplotlib2tikz v0.6.13.
\begin{tikzpicture}

\definecolor{color0}{rgb}{0.12156862745098,0.466666666666667,0.705882352941177}

\begin{axis}[
xmin=-210, xmax=4410,
ymin=0, ymax=226.8,
tick align=outside,
tick pos=left,
x grid style={lightgray!92.026143790849673!black},
y grid style={lightgray!92.026143790849673!black},
xlabel={Overlap length},
title={Histogram of overlap lengths},
]
\draw[fill=color0,draw opacity=0] (axis cs:0,0) rectangle (axis cs:100,94);
\draw[fill=color0,draw opacity=0] (axis cs:100,0) rectangle (axis cs:200,216);
\draw[fill=color0,draw opacity=0] (axis cs:200,0) rectangle (axis cs:300,95);
\draw[fill=color0,draw opacity=0] (axis cs:300,0) rectangle (axis cs:400,34);
\draw[fill=color0,draw opacity=0] (axis cs:400,0) rectangle (axis cs:500,46);
\draw[fill=color0,draw opacity=0] (axis cs:500,0) rectangle (axis cs:600,57);
\draw[fill=color0,draw opacity=0] (axis cs:600,0) rectangle (axis cs:700,113);
\draw[fill=color0,draw opacity=0] (axis cs:700,0) rectangle (axis cs:800,24);
\draw[fill=color0,draw opacity=0] (axis cs:800,0) rectangle (axis cs:900,17);
\draw[fill=color0,draw opacity=0] (axis cs:900,0) rectangle (axis cs:1000,17);
\draw[fill=color0,draw opacity=0] (axis cs:1000,0) rectangle (axis cs:1100,19);
\draw[fill=color0,draw opacity=0] (axis cs:1100,0) rectangle (axis cs:1200,13);
\draw[fill=color0,draw opacity=0] (axis cs:1200,0) rectangle (axis cs:1300,8);
\draw[fill=color0,draw opacity=0] (axis cs:1300,0) rectangle (axis cs:1400,9);
\draw[fill=color0,draw opacity=0] (axis cs:1400,0) rectangle (axis cs:1500,14);
\draw[fill=color0,draw opacity=0] (axis cs:1500,0) rectangle (axis cs:1600,6);
\draw[fill=color0,draw opacity=0] (axis cs:1600,0) rectangle (axis cs:1700,8);
\draw[fill=color0,draw opacity=0] (axis cs:1700,0) rectangle (axis cs:1800,19);
\draw[fill=color0,draw opacity=0] (axis cs:1800,0) rectangle (axis cs:1900,10);
\draw[fill=color0,draw opacity=0] (axis cs:1900,0) rectangle (axis cs:2000,4);
\draw[fill=color0,draw opacity=0] (axis cs:2000,0) rectangle (axis cs:2100,10);
\draw[fill=color0,draw opacity=0] (axis cs:2100,0) rectangle (axis cs:2200,9);
\draw[fill=color0,draw opacity=0] (axis cs:2200,0) rectangle (axis cs:2300,2);
\draw[fill=color0,draw opacity=0] (axis cs:2300,0) rectangle (axis cs:2400,5);
\draw[fill=color0,draw opacity=0] (axis cs:2400,0) rectangle (axis cs:2500,13);
\draw[fill=color0,draw opacity=0] (axis cs:2500,0) rectangle (axis cs:2600,9);
\draw[fill=color0,draw opacity=0] (axis cs:2600,0) rectangle (axis cs:2700,8);
\draw[fill=color0,draw opacity=0] (axis cs:2700,0) rectangle (axis cs:2800,15);
\draw[fill=color0,draw opacity=0] (axis cs:2800,0) rectangle (axis cs:2900,8);
\draw[fill=color0,draw opacity=0] (axis cs:2900,0) rectangle (axis cs:3000,4);
\draw[fill=color0,draw opacity=0] (axis cs:3000,0) rectangle (axis cs:3100,11);
\draw[fill=color0,draw opacity=0] (axis cs:3100,0) rectangle (axis cs:3200,16);
\draw[fill=color0,draw opacity=0] (axis cs:3200,0) rectangle (axis cs:3300,8);
\draw[fill=color0,draw opacity=0] (axis cs:3300,0) rectangle (axis cs:3400,3);
\draw[fill=color0,draw opacity=0] (axis cs:3400,0) rectangle (axis cs:3500,6);
\draw[fill=color0,draw opacity=0] (axis cs:3500,0) rectangle (axis cs:3600,10);
\draw[fill=color0,draw opacity=0] (axis cs:3600,0) rectangle (axis cs:3700,8);
\draw[fill=color0,draw opacity=0] (axis cs:3700,0) rectangle (axis cs:3800,6);
\draw[fill=color0,draw opacity=0] (axis cs:3800,0) rectangle (axis cs:3900,7);
\draw[fill=color0,draw opacity=0] (axis cs:3900,0) rectangle (axis cs:4000,3);
\draw[fill=color0,draw opacity=0] (axis cs:4000,0) rectangle (axis cs:4100,14);
\draw[fill=color0,draw opacity=0] (axis cs:4100,0) rectangle (axis cs:4200,6);
\end{axis}

\end{tikzpicture}}
    \caption{Histogram of the lengths of highly correlated regions as identified by our analysis.}
    \label{fig:histogram}
    \end{minipage}
    \begin{minipage}{.6\textwidth}
    \centering
    \resizebox{7cm}{!}{{\input{./35}}}
    \caption{Time series that repeats itself.}
    \label{fig:match_itself}
    \end{minipage}
\end{figure}

\begin{table}[h]
    \centering
    \def\arraystretch{1.5}
    \caption{Number of time series in categories T1-T4. Table shows how many time series were assigned to one of the categories T1-T4, and the performance of forecasts when using these similarities.}
	\label{tab:t_cat}
\begin{tabular}{|l|l|l|l|l|l|}
\hline
\textbf{}                 & \textbf{T1} & \textbf{T2} & \textbf{T3} & \textbf{T4} & \textbf{All} \\ \hline
\rowcolor[HTML]{EFEFEF} 
\textbf{Time series (\#)} & 12          & 202         & 23          & 767         & 1004         \\ \hline
\textbf{MASE}             & 1.615       & 0.077       & 0.702       & 0.471       & 0.410        \\ \hline
\rowcolor[HTML]{EFEFEF} 
\textbf{sMAPE}            & 1.418       & 0.084       & 0.540       & 0.424       & 0.370        \\ \hline
\textbf{OWA}              & 0.482       & 0.030       & 0.202       & 0.166       & 0.144        \\ \hline
\end{tabular}
\end{table}

The above results raise the question of whether the success of the correlator in the M4 competition could be fully explained by leakage in forecasting by using information from the future. 
Our additional analysis revealed that the correlator used future data in only 26\% of the cases where it was applied. Allowing the correlator to use only past values resulted in OWA of 0.895 and 0.917 with correlation thresholds of 0.999 and 0.9999 respectively. These results lead us to the hypothesis that possibly there are further data quality issues, including duplications of data within one time series explaining some of the correlations of type T1, rearrangement errors of time series segments explaining some of T2 cases, and wrong starting dates explaining cases in T4.

Importantly, the evidence we presented here strongly suggests that the success of the correlator in forecasting future from the past was only due to data quality issues in the M4 data, most probably going back to the issues in the databases where the M4 data originate from.

\begin{figure}[h]
    \hspace{-1cm}
    \begin{minipage}{.6\textwidth}
    \centering
	\resizebox{7cm}{!}{{\input{./T2_163}}}
	
	\resizebox{7cm}{!}{{\input{./T2_446}}}
    \caption{Time series such that the beginning of one is correlated with the ending of the other and vice versa.}
    \label{fig:t2_example}
    \end{minipage}
    \begin{minipage}{.6\textwidth}
    \centering
	\resizebox{7cm}{!}{{{% This file was created by matplotlib2tikz v0.6.13.
\begin{tikzpicture}

\definecolor{color0}{rgb}{0.917647058823529,0.917647058823529,0.949019607843137}
\definecolor{color1}{rgb}{0.298039215686275,0.447058823529412,0.690196078431373}
\definecolor{color2}{rgb}{0.333333333333333,0.658823529411765,0.407843137254902}

\begin{axis}[
xmin=-11.85, xmax=248.85,
ymin=-526.95, ymax=11065.95,
tick align=outside,
tick pos=left,
legend cell align={left},
legend entries={{D45},{D46}},
legend style={at={(0.97,0.97)}, anchor=north east, draw=none},
width=15cm,
height=5cm,
title={Time series D45, correlated with D46},
]
\addplot [thick, color1]
table {%
0 997.8
1 994.6
2 997.2
3 1002
4 996.5
5 997.1
6 996.1
7 999.5
8 1000.8
9 1007
10 1011.7
11 1006.4
12 1006.4
13 998.1
14 978
15 984.8
16 979.4
17 975.9
18 989
19 982.6
20 992
21 992.8
22 987.4
23 982.7
24 993.1
25 985.5
26 968
27 981.5
28 983.8
29 963
30 976.1
31 984.6
32 979.3
33 978.1
34 997.3
35 995.7
36 995.4
37 1009.2
38 1006.1
39 1004.7
40 1003.2
41 1016.1
42 1016.1
43 1005.8
44 999.9
45 1004.8
46 1000.1
47 998.4
48 990.9
49 991.5
50 985.7
51 986.8
52 987.6
53 1001.1
54 987
55 979.7
56 967.9
57 966.2
58 973.1
59 992.2
60 997.2
61 987
62 980.3
63 973.7
64 957.7
65 955.1
66 960.5
67 965
68 964.8
69 999
70 1007
71 1003
72 993.4
73 1008.7
74 1008.7
75 1012.4
76 1000.5
77 1024
78 1029.8
79 1032.9
80 1033.1
81 1037.5
82 1033.8
83 1037
84 1044.3
85 1043.3
86 1048.4
87 1051.9
88 1051.1
89 1053.9
90 1031
91 1026.8
92 1046.7
93 1028.5
94 1032.7
95 1023.3
96 1016.7
97 1019.8
98 1003.4
99 990.4
100 1003.4
101 1001.5
102 989.9
103 982
104 982.3
105 984.7
106 993.3
107 993.2
108 993.2
109 987.9
110 955.9
111 932.4
112 976.9
113 960.5
114 964.9
115 960.6
116 966.2
117 962.8
118 959.4
119 967.9
120 983.8
121 991.5
122 991.6
123 997.4
124 995.8
125 1001.3
126 996.1
127 1003
128 1014.9
129 1016.1
130 1017
131 1020.2
132 1027.3
133 1027.3
134 1020.2
135 1020.2
136 1020.2
137 1017.7
138 1029
139 1029
140 1029.4
141 1021.5
142 1021
143 1028.8
144 1036.9
145 1037.6
146 1044.1
147 1045.5
148 1036
149 1041
150 1032.6
151 1021.9
152 1015.4
153 1026.5
154 1031.7
155 1030.2
156 1010.8
157 1012
158 1020.4
159 1009.6
160 1025.5
161 1029.4
162 1040.3
163 1046.3
164 1046
165 1060
166 1065
167 1065.9
168 1068
169 1073.7
170 1067.6
171 1072
172 1075.7
173 1072
174 1066
175 1070
176 1071
177 1071.1
178 1068
179 1060.3
180 1071
181 1071
182 1063
183 1067
184 1065.2
185 1057.7
186 1055
187 1074.6
188 1078
189 1081
190 1081
191 1099
192 1098
193 1104
194 1106.2
195 1110
196 1100.4
197 1107
198 1106
199 1102
200 1110.3
201 1110.9
202 1115.3
203 1113.7
204 1118.2
205 1115.3
206 1119.8
207 1118.8
208 1117
209 1119.6
210 1113.4
211 1107.5
212 1115.6
213 1118.7
214 1125.8
215 1116.8
216 1117.8
217 1117.9
218 1119.2
219 1122
220 1115.5
221 1121
222 1124.4
223 1120.1
224 1103.2
225 1095.7
226 1083.8
227 1092
228 1092
229 1086
230 1079.5
231 1071.7
232 1069.1
233 1083.6
234 1076
235 1084.6
236 1104
237 1110.6
};
\addplot [thick, red, dashed]
table {%
0 9978
1 9946
2 9972
3 10020
4 9965
5 9971
6 9961
7 9995
8 10008
9 10070
10 10117
11 10064
12 10064
13 9981
14 9780
15 9848
16 9794
17 9759
18 9890
19 9826
20 9920
21 9928
22 9874
23 9827
24 9931
25 9855
26 9680
27 9815
28 9838
29 9630
30 9761
31 9846
32 9793
33 9781
34 9973
35 9957
36 9954
37 10092
38 10061
39 10047
40 10032
41 10161
42 10161
43 10058
44 9999
45 10048
46 10001
47 9984
48 9909
49 9915
50 9857
51 9868
52 9876
53 10011
54 9870
55 9797
56 9679
57 9662
58 9731
59 9922
60 9972
61 9870
62 9803
63 9737
64 9577
65 9551
66 9605
67 9650
68 9648
69 9990
70 10070
71 10030
72 9934
73 10087
74 10087
75 10124
76 10005
77 10240
78 10298
79 10329
80 10331
81 10375
82 10338
83 10370
84 10443
85 10433
86 10484
87 10519
88 10511
89 10539
90 10310
91 10268
92 10467
93 10285
94 10327
95 10233
96 10167
97 10198
98 10034
99 9904
100 10034
101 10015
102 9899
103 9820
104 9823
105 9847
106 9933
107 9932
108 9932
109 9785
110 9465
};
\end{axis}

\end{tikzpicture}}}}
	
	\resizebox{7cm}{!}{{{% This file was created by matplotlib2tikz v0.6.13.
\begin{tikzpicture}

\definecolor{color0}{rgb}{0.917647058823529,0.917647058823529,0.949019607843137}
\definecolor{color1}{rgb}{0.298039215686275,0.447058823529412,0.690196078431373}
\definecolor{color2}{rgb}{0.333333333333333,0.658823529411765,0.407843137254902}

\begin{axis}[
xmin=-11.85, xmax=248.85,
tick align=outside,
tick pos=left,
legend cell align={left},
legend entries={{D45 standardized},{D46 standardized}},
legend style={at={(0.97,0.03)}, anchor=south east, draw=none},
width=15cm,
height=5cm,
xlabel={Days},
]
\addplot [thick, color1]
table {%
0 -0.0556569916304715
1 -0.199853661412777
2 -0.0826938672146503
3 0.133601137458811
4 -0.114236888729532
5 -0.0872000131453485
6 -0.13226147245232
7 0.0209474891913822
8 0.0795273862904432
9 0.358908433993669
10 0.570697292736437
11 0.331871558409485
12 0.331871558409485
13 -0.0421385538383769
14 -0.947873885908506
15 -0.641455962621102
16 -0.884787842878747
17 -1.04250295045315
18 -0.452197833531819
19 -0.740591173096436
20 -0.317013455610904
21 -0.280964288165329
22 -0.524296168422975
23 -0.736085027165738
24 -0.267445850373235
25 -0.60991294110622
26 -1.39848847897822
27 -0.790158778334106
28 -0.686517421928073
29 -1.62379577551308
30 -1.03349065859175
31 -0.650468254482493
32 -0.889293988809445
33 -0.943367739977808
34 -0.0781877212839573
35 -0.150286056175108
36 -0.163804493967202
37 0.458043644469008
38 0.318353120617396
39 0.255267077587636
40 0.187674888627179
41 0.768967713687111
42 0.768967713687111
43 0.304834682825301
44 0.0389720729141698
45 0.259773223518329
46 0.0479843647755662
47 -0.0286201160462875
48 -0.366581060848574
49 -0.33954418526439
50 -0.600900649244823
51 -0.551333044007159
52 -0.515283876561578
53 0.0930458240825378
54 -0.542320752145762
55 -0.871269405086653
56 -1.40299462490892
57 -1.47959910573077
58 -1.16867503651267
59 -0.308001163749508
60 -0.0826938672146503
61 -0.542320752145762
62 -0.844232529502474
63 -1.14163816092848
64 -1.86262150984003
65 -1.97978130403815
66 -1.73644942378051
67 -1.53367285689914
68 -1.54268514876053
69 -0.00158324046210355
70 0.358908433993669
71 0.178662596765783
72 -0.253927412581145
73 0.435512914815523
74 0.435512914815523
75 0.602240314251314
76 0.0660089484983538
77 1.12495324221219
78 1.38630970619262
79 1.52600023004424
80 1.53501252190562
81 1.7332829428563
82 1.5665555434205
83 1.71075221320282
84 2.03970086614371
85 1.99463940683673
86 2.2244528493023
87 2.3821679568767
88 2.34611878943111
89 2.47229087549064
90 1.44038345736099
91 1.2511253282717
92 2.14784836848044
93 1.32772980909356
94 1.51698793818284
95 1.0934102206973
96 0.796004589271295
97 0.935695113122903
98 0.19668718048857
99 -0.38911179050206
100 0.19668718048857
101 0.111070407805325
102 -0.411642520155546
103 -0.76762804868062
104 -0.754109610888531
105 -0.645962108551795
106 -0.258433558511844
107 -0.262939704442537
108 -0.262939704442537
109 -0.501765438769489
110 -1.94373213659258
111 -3.00267643030641
112 -0.997441491146176
113 -1.73644942378051
114 -1.53817900282983
115 -1.73194327784981
116 -1.47959910573077
117 -1.63280806737448
118 -1.78601702901818
119 -1.40299462490892
120 -0.686517421928073
121 -0.33954418526439
122 -0.335038039333692
123 -0.0736815753532591
124 -0.145779910244415
125 0.102058115943929
126 -0.13226147245232
127 0.178662596765783
128 0.714893962518743
129 0.768967713687111
130 0.809523027063385
131 0.953719696845696
132 1.27365605792519
133 1.27365605792519
134 0.953719696845696
135 0.953719696845696
136 0.953719696845696
137 0.841066048578267
138 1.35026053874704
139 1.35026053874704
140 1.36828512246984
141 1.01229959394476
142 0.989768864291271
143 1.34124824688565
144 1.70624606727212
145 1.73778908878699
146 2.03068857428231
147 2.09377461731207
148 1.66569075389584
149 1.8909980504307
150 1.51248179225214
151 1.03032417766754
152 0.737424692172229
153 1.23760689047961
154 1.47192647887587
155 1.40433428991541
156 0.530141979360159
157 0.584215730528527
158 0.962731988707087
159 0.476068228191796
160 1.19254543117264
161 1.36828512246984
162 1.85945502891582
163 2.12982378475765
164 2.11630534696556
165 2.74716577726316
166 2.97247307379802
167 3.0130283871743
168 3.10765745171893
169 3.36450776976867
170 3.08963286799614
171 3.28790328894682
172 3.45463068838262
173 3.28790328894682
174 3.01753453310499
175 3.19778037033288
176 3.24284182963985
177 3.24734797557054
178 3.10765745171893
179 2.76068421505525
180 3.24284182963985
181 3.24284182963985
182 2.88235015518408
183 3.06259599241196
184 2.98148536565942
185 2.64352442085713
186 2.5218584807283
187 3.40506308314494
188 3.55827204478865
189 3.69345642270956
190 3.69345642270956
191 4.50456269023505
192 4.45950123092808
193 4.72986998676991
194 4.82900519724525
195 5.00023874261174
196 4.56764873326482
197 4.86505436469082
198 4.81999290538385
199 4.63974706815597
200 5.01375718040383
201 5.04079405598802
202 5.23906447693869
203 5.16696614204754
204 5.36974270892891
205 5.23906447693869
206 5.44184104382006
207 5.39677958451309
208 5.31566895776054
209 5.43282875195866
210 5.15344770425545
211 4.88758509434431
212 5.25258291473078
213 5.39227343858239
214 5.71220979966189
215 5.30665666589914
216 5.35171812520612
217 5.35622427113682
218 5.41480416823588
219 5.5409762542954
220 5.24807676880008
221 5.49591479498843
222 5.64912375663213
223 5.45535948161215
224 4.69382081932434
225 4.35585987452205
226 3.81962850876908
227 4.18913247508625
228 4.18913247508625
229 3.91876371924442
230 3.62586423374911
231 3.27438485115473
232 3.1572250569566
233 3.81061621690769
234 3.46814912617471
235 3.85567767621466
236 4.72986998676991
237 5.02727561819592
};
\addplot [thick, red, dashed]
table {%
0 -0.0475089969066752
1 -0.190156466908803
2 -0.0742553975320742
3 0.139715807471118
4 -0.10545953159504
5 -0.0787131309696407
6 -0.123290465345306
7 0.0282724715319555
8 0.0862230062203201
9 0.362602479349444
10 0.57211595091507
11 0.335856078724045
12 0.335856078724045
13 -0.0341357965939756
14 -0.930140217544844
15 -0.627014343790321
16 -0.867731949418913
17 -1.02375261973374
18 -0.439789539412528
19 -0.725084479416784
20 -0.306057536285533
21 -0.270395668785001
22 -0.511113274413592
23 -0.720626745979218
24 -0.257022468472301
25 -0.595810209727356
26 -1.37591356130149
27 -0.774119547230016
28 -0.671591678165986
29 -1.59880023317982
30 -1.01483715285861
31 -0.635929810665454
32 -0.872189682856479
33 -0.925682484107277
34 -0.0697976640945077
35 -0.141121399095572
36 -0.154494599408271
37 0.460672614975907
38 0.322482878411345
39 0.260074610285414
40 0.193208608721916
41 0.768256222167996
42 0.768256222167996
43 0.309109678098645
44 0.0461034052822215
45 0.26453234372298
46 0.0550188721573545
47 -0.0207625962812761
48 -0.355092604098764
49 -0.328346203473365
50 -0.586894742852223
51 -0.537859675038991
52 -0.502197807538459
53 0.0995962065330196
54 -0.528944208163858
55 -0.854358749106213
56 -1.38037129473906
57 -1.45615276317769
58 -1.1485691559856
59 -0.2971420694104
60 -0.0742553975320742
61 -0.528944208163858
62 -0.827612348480814
63 -1.1218227553602
64 -1.83506010537084
65 -1.95096117474757
66 -1.71024356911898
67 -1.50964556442849
68 -1.51856103130362
69 0.00598380434412293
70 0.362602479349444
71 0.184293141846783
72 -0.243649268159602
73 0.438383947788074
74 0.438383947788074
75 0.603320084978035
76 0.0728498059076206
77 1.12041716373575
78 1.37896570311461
79 1.51715543967917
80 1.5260709065543
81 1.72221117780723
82 1.55727504061727
83 1.6999225106194
84 2.02533705156175
85 1.98075971718609
86 2.20810412250198
87 2.36412479281681
88 2.32846292531627
89 2.45327946156814
90 1.43245850436541
91 1.24523369998761
92 2.13232265406335
93 1.32101516842624
94 1.50823997280404
95 1.08921302967278
96 0.795002622793395
97 0.933192359357957
98 0.202124075597049
99 -0.377381271286597
100 0.202124075597049
101 0.117427140283286
102 -0.399669938474429
103 -0.751830880042183
104 -0.738457679729484
105 -0.631472077227888
106 -0.248107001597168
107 -0.252564735034735
108 -0.252564735034735
109 -0.907851550357011
110 -2.33432625037829
};
\end{axis}

\end{tikzpicture}}}}
    \caption{Time series that match after standardising.}
    \label{fig:scaled_match}
    \end{minipage}
\end{figure}

\begin{figure}[h]
    \hspace{-1cm}\begin{minipage}{.6\textwidth}
    \centering
    \resizebox{7cm}{!}{{\input{./102}}}
    \caption{Time series that are shifted versions of one another.}
    \label{fig:match_shift}
    \end{minipage}
    \begin{minipage}{.6\textwidth}
    	\centering
	\resizebox{7cm}{!}{{\input{./rearrange3}}}
	
    \resizebox{7cm}{!}{{\input{./rearrange2}}}
	\caption{Time series that can be split into large segments and rearranged into one another.}
	\label{fig:correlated_new}
	\end{minipage}
\end{figure}

\section{Conclusions}

In this article, we have described our method to obtain forecasts for Daily time series within the M4 competition. We used an ensemble of five statistical forecasting methods and a method that we refer to as the correlator. The correlator was applied on a time series only if its last 14 days were found to correlate strongly with another time series. 
Our retrospective analysis using the full time series published by the M4 organisers after the competition demonstrated that the correlator was responsible for most of our gains over the naive constant forecasting method. We found that in 26\% of cases where the correlator was applied it actually used data from the future. However, as the actual starting dates were only published by the M4 organisers after the competition we had no way of filtering those cases out in our submission. To avoid this problem we suggest providing the starting dates explicitly in future competitions. 
In this analysis we have revealed some potential data quality issues, including potential duplications and rearrangement errors within particular time series and potential errors in starting dates. We cannot rule out the possibility that success of the correlator in forecasting the future from the past was only due to data quality issues in the time series.

\section*{Acknowledgements}

\anonymize{Our submission to the M4 time series forecasting competition was prepared during a seminar dedicated to M4 at the University of Tartu. The authors would like to thank all other participants of the seminar for useful discussions: %\todo[inline]{list all non-author participants. Novin: Done} % taken from https://goo.gl/Vsyw9g
Alina Vorontseva, Anton Potapchuk, Aytaj Aghabayli, Basar Turgut, Diana Grygorian, Gunay Abdullayeva, Jayasinghe Arachchilage Sriyal Himesh Jayasinghe, Joonas Puura, Kaur Karus, Maksym Semikin, Markus Loide, Martin Liivak, Martin Valgur, Mikhail	Papkov, Märten Veskimäe, Olha Kaminska, Prabhant Singh, Saumitra Bagchi and Yevheniia Kryvenko.}

%\section*{References}
\bibliography{article}

\appendix
\section{Implementation of the algorithm}\label{appendix_repo}
The code for producing the forecasts using our method can be found in Github repository\footnote{\url{\anonymize{https://github.com/antiingel/correlator}}
}. The detailed instruction on how to run the code is in \texttt{README.md} file in the repository.

\section{ARIMA and ETS models}\label{sec:arima}

Two ARIMA models~\citep{box2015time} and an ETS model~\citep{ETS} were fitted using corresponding functions in the \texttt{forecast} package version 8.3~\citep{forecast2} in the \texttt{R} software version 3.2.4. ETS was fitted using function \texttt{ets} with the default parameter settings. ARIMA models were fitted using \texttt{auto.arima} function. In the following we discuss the differences between the two ARIMA models.

The first ARIMA model was obtained using the stepwise algorithm~\citep{forecast2} with the default settings except the starting values for the order of the autoregressive model and the moving average model which were set to 0. We refer to this method as ARIMA 1.
    
The second model was obtained using a non-stepwise algorithm that searches through all the (S)ARIMA models with the maximum allowed order for the model set to 8. The fitting of ARIMA models was done using maximum likelihood estimation instead of using approximation with conditional sum of squares as in ARIMA 1 method. We refer to this method as ARIMA 2. 

Our retrospective analysis showed that using only one ARIMA model instead of two in the ensemble gave better results as outlined in Section~\ref{sec:results}. Therefore, since ARIMA 2 is much more computationally expensive, using only ARIMA 1 would have been beneficial.

\section{Custom method}\label{sec:fearless}

Our custom time series forecasting method\footnote{Originally named \emph{fearless} in our submission} is inspired by \textit{Forecasting with Decomposition} from \cite{hyndman2014forecasting}. Our method differs from the method described by \cite{hyndman2014forecasting} in mainly two ways. First, we use the classical decomposition~\citep{NBERmaca31-1} instead of STL decomposition~\citep{STL_decomp} and secondly, we use either linear extrapolation~\citep{regression} or past values from the decomposition for forecasting trend and residuals. 

In more detail, classical additive decomposition is performed on the time series with frequency set to 2 instead of 1 as in the Daily dataset. This means that a moving average filter with weights $(0.25, 0.5, 0.25)$ is applied to the time series to obtain the trend and then the seasonal component with period 2 is calculated from that, see \cite{hyndman2014forecasting} for details. However, in most cases the seasonal component is very small relative to other components since the time series does not actually have frequency 2. Therefore, setting frequency to 2 mainly serves to reduce noise in the time series by applying moving average filter. For decomposition \texttt{seasonal\_decompose} function in \texttt{StatsModel} package version 0.8.0 \citep{seabold2010statsmodels} was used.

To predict the trend and the residuals, two approaches for the trend and for the residuals are considered. In total, all four combinations of these approaches are evaluated on the internal validation set and the approach that performs the best according to MASE evaluation measure~\citep{MASE} is used on the actual dataset. One considered approach is linear extrapolation from the last 14 values, which was performed using \texttt{curve\_fit} from \texttt{Scipy} package version 1.0.1 \citep{scipy}. The other approach is using the previous 14 values of the corresponding component as the forecasts. 

Linear functions were chosen for the extrapolation, because the method was mainly developed for Yearly data, in which training samples were limited and in many cases the trend was not changing dramatically. Therefore, the poor performance on Daily dataset was unsurprising. For Yearly dataset our custom method achieved OWA of 0.967. We decided to include the custom method in the Daily ensemble to introduce some variability.

Seasonal component is predicted as seasonal naive, which means that the forecast for current period is the last period's corresponding value. The final forecast is the sum of the predicted trend, residual and seasonal component. In our original submission, this method was used for all Daily time series except for time series D3160 which had incorrect forecast due to a bug in our code.

\end{document}